\documentclass[12pt,a4paper]{cibb}
\usepackage[table]{xcolor}
\usepackage{subfigure,graphicx}
\usepackage{multirow}
\usepackage{amsmath,amsfonts,latexsym,amssymb,euscript,xr}

\title{\large $\ $\\ \bf Knowledge Graph-based Neurodegenerative Diseases and Diet Relationship Discovery}

\author{ Yi Nian$^{1}$ , Jingcheng Du$^{1}$, Larry Bu$^{2}$, Fang Li$^{1}$, Xinyue Hu$^{1}$, Yuji Zhang$^{2,3}$, Cui Tao$^{*1}$}

\address{$\ $\\$^1$ School of Biomedical Informatics, University of Texas Health Science Center at Houston, Houston, TX, USA \\
\bigskip
$^2$  University of Maryland School of Medicine, Baltimore MD USA\\
\bigskip
$^3$ The University of Maryland Greenebaum Comprehensive Cancer Center, Baltimore MD USA
\bigskip\\

$^*$corresponding author (email:cui.tao@uth.tmc.edu)
}

\abstract{knowledge graph, neurodegenerative disease, diet.
\\[17pt]
{\bf Abstract.} To date, there are no effective treatments for most neurodegenerative diseases. However, certain foods may be associated with these diseases and bring an opportunity to prevent or delay neurodegenerative progression. Our objective is to construct a knowledge graph for neurodegenerative diseases using literature mining to study their relations with diet. We collected biomedical annotations (Disease, Chemical, Gene, Species, SNP\&Mutation) in the abstracts from 4,300 publications relevant to both neurodegenerative diseases and diet using PubTator, an NIH-supported tool that can extract biomedical concepts from literature. A knowledge graph was created from these annotations. Graph embeddings were then trained with the node2vec algorithm to support potential concept clustering and similar concept identification. We found several food-related species and chemicals that might come from diet and have an impact on neurodegenerative diseases.}

\begin{document}
\thispagestyle{myheadings}
\pagestyle{myheadings}
\markright{\tt Proceedings of CIBB 2021}

\section{\bf Scientific Background}

Neurodegenerative diseases are a heterogeneous group of disorders that are characterized by the progressive degeneration of the structure and function of the central nervous system or peripheral nervous system.[1] Common neurodegenerative diseases, such as Alzheimer's disease and Parkinson's disease, are usually incurable and difficult to stop and irreversible. Neurodegenerative disease affects humans in different activities, such as balance, movement, talking, and breathing. Studies have indicated that diets could be related to prevent or delay neurodegenerative diseases and cognitive decline [2]. However, further research is needed to better understand the backend mechanisms and to reveal the potential interactions with clinical and pharmacokinetic factors.

The objective of this paper is to study potential relations between neurodegenerative diseases and diet using a knowledge graph-based approach. The concept of knowledge graph originated from Google and was used to enhance information retrieval from different sources. In this paper, we encode biomedical concepts and their rich relations into a network (knowledge graph) through literature mining [3]. Literature Mining is a data mining technique that identifies the entities such as genes, diseases, and chemicals from literature, discovers global trends, and facilitates hypothesis generation based on existing knowledge. Literature mining enables researchers to study a massive amount of literature quickly and reveal hidden relations between entities that might be hard to discover by manual analysis. In this paper, we introduce a biomedical knowledge graph that specifically focuses on neurodegenerative diseases and diet. Since foods may consist of certain chemicals and species, this knowledge graph could be used to study and discover underlying relations between diet and neurodegenerative diseases.

\section{\bf Materials and Methods}

We first retrieved the abstracts that were related to neurodegenerative diseases and food/diet from PubMed [4]. Biomedical entities in the abstracts were then extracted using PubTator [5]. In this paper, relations between entities were determined by co-occurrence, based on which the knowledge graph was constructed. Finally, we generated node embeddings and analyzed two representative neurodegenerative diseases based on the cluster of embeddings. A visualization of our knowledge graph is provided using Neo4j. An overview of this pipeline is illustrated in Figure 1.

\begin{figure}[h]
\vspace{3mm}
 \begin{center}
 \includegraphics[width=12cm]{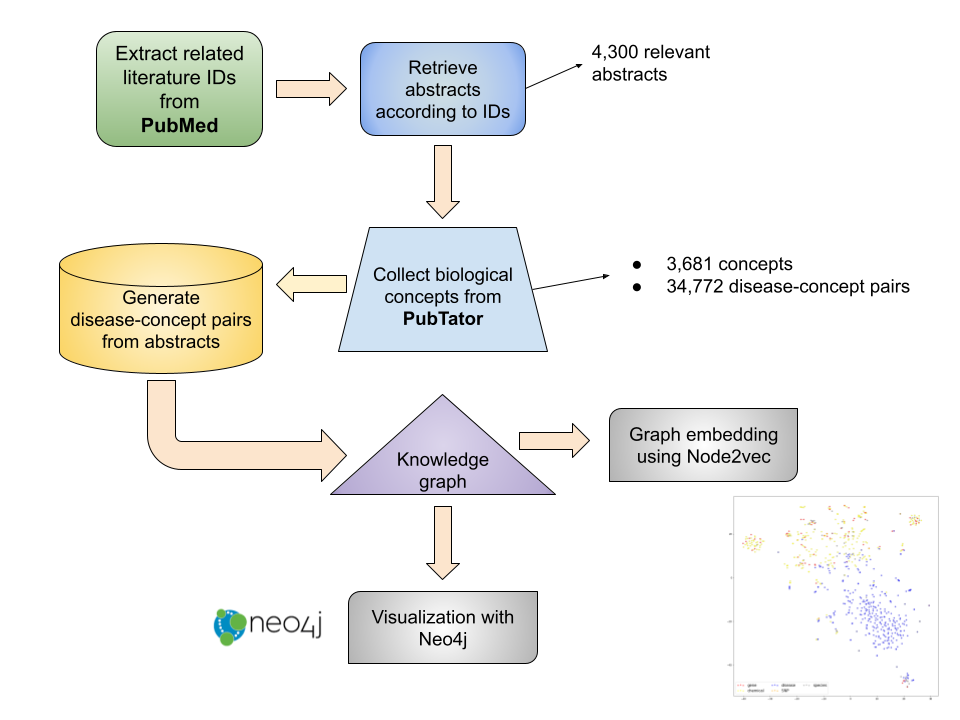}
\caption {Overview of pipeline.\label{cibb-fig}}
 \end{center}
\vspace{-8mm}
\end{figure}

\subsection{\bf \it Search of Related Papers}

To retrieve abstracts that are relevant to both neurodegenerative diseases and diet, we used the query “(Alzheimer's disease OR Parkinson’s disease OR Prion disease OR Huntington disease OR neurodegenerative disease) AND (eat OR diet OR food)”, to search the PubMed database and we only used the abstracts of these papers. PubMed is a public database that comprises more than 32 million citations for biomedical literature from MEDLINE, life science journals, and online books [4]. The abstracts collected were from different types of studies: Randomized Controlled Trials, Clinical Trials, and Meta-Analyses. The publication dates of these studies ranged from 1975 to 2020. 

\subsection{\bf \it  Biomedical Concept Extraction}

PubTator is a web-based tool developed by NCBI that provides automatic annotations of biomedical concepts such as genes, chemicals and diseases. Using Pubtator, we extracted different biomedical entities from the relevant abstracts. We further classified these entities into five concept categories: Disease, Chemical, Species, Gene, and SNP\&Mutation (including DNA and protein mutations). We assumed that co-occurrence in the same abstract indicates a certain relationship might exists between the two entities. For every occurrence of Disease in one abstract, we linked it with all other co-occurring concepts. We iterated this process for all abstracts and these pairs are used to construct our knowledge graph. In the knowledge graph, we created a node for each biomedical concept. For each pair of nodes, we also incorporated information such as their occurrence frequency and their source literature into the edge.

\subsection{\bf \it Network embeddings representation learning}

We leveraged Neo4j, a well-acknowledged graph database management system to construct a knowledge graph for all Disease-concept relationships. Figure 2 is the visualization of this knowledge graph where different colors represent different types of biomedical concepts. To quantitatively compare the relationship between different biomedical concepts, we leveraged node2vec [7] to map each graph node into a fix-length vector that maximizes the likelihood of preserving network neighborhoods of nodes. Specifically, node2vec generates fixed-length random walks using different nodes as initial nodes and feeds them into the word2vec algorithm to get the numerical representation. Here we used a random walk length of 10 and the number of dimensions for representations is 100. Moreover, we also used the occurrence frequency of each Disease-related pair as the weight between every two nodes. Node2vec calculates the transition probabilities between nodes using this weight and the probability is leveraged to generate random walks. To visualize the embedding space, we further applied the t-distributed stochastic neighbor embedding (t-SNE) [8] approach to reduce the dimension from 100 to 2 while still preserving the local structure. Specifically, t-SNE:

1. Converts the shortest distance between points into a probability distribution of similar points.

2. Calculates a similar pairwise conditional probability in low dimensional space using a heavy-tailed t-distribution,

3. Minimizes the sum of the difference in conditional probabilities using Kullback-Leibler divergences between step 1 and step 2 with gradient descent.

\begin{figure}[h]
\vspace{3mm}
 \begin{center}
 \includegraphics[width=12cm]{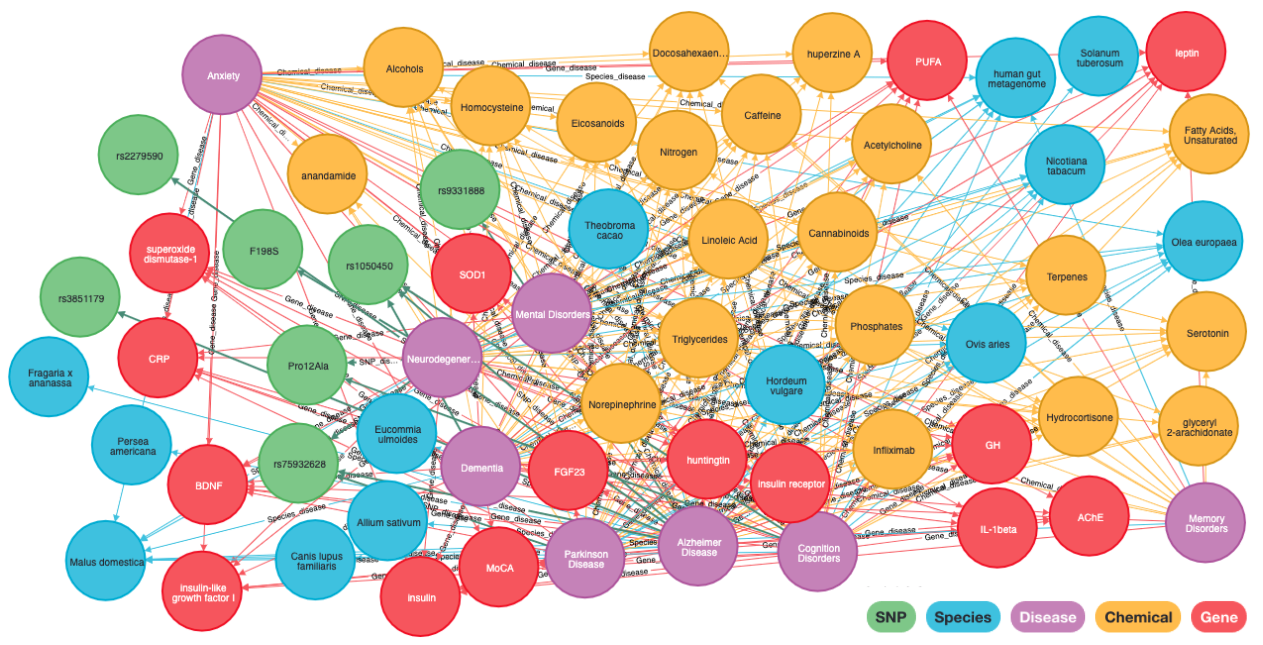}
\caption { Neo4j knowledge graph visualization.}
 \end{center}
\vspace{-8mm}
\end{figure}

\begin{figure}[hbt!]
\vspace{3mm}
 \begin{center}
 \includegraphics[width=11cm]{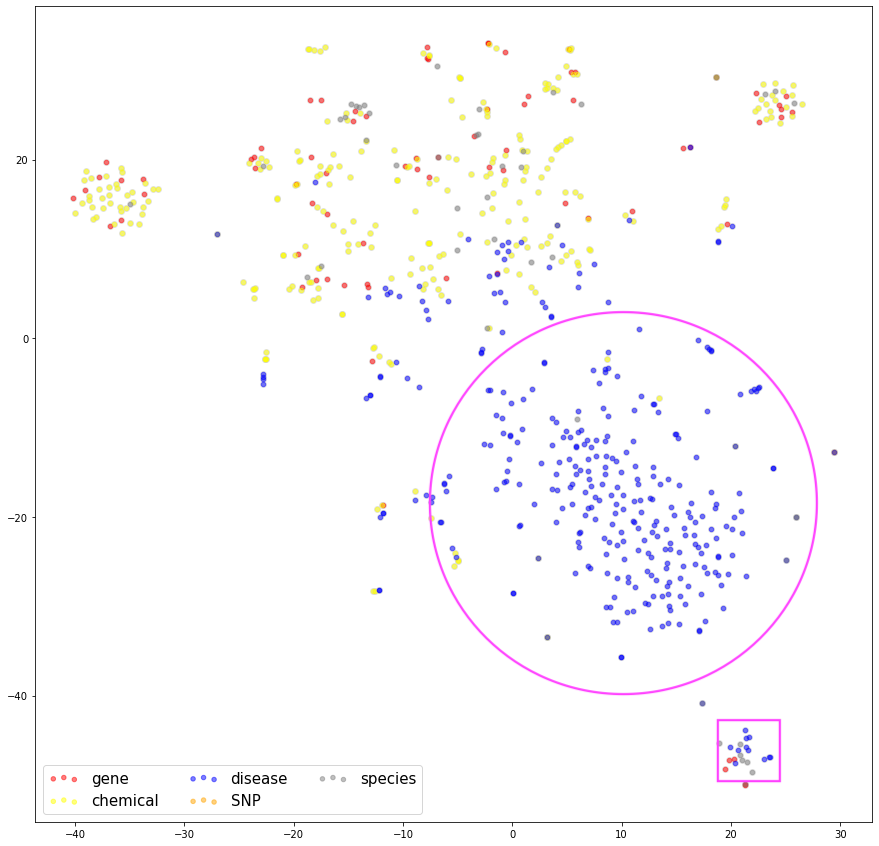}
\caption {T-SNE visualization of Node2vec embeddings.}
 \end{center}
\vspace{-8mm}
\end{figure}

A graphical result using t-SNE from node embedding data is shown in Figure 3. Clear clusters of diseases can be seen in the lower part of Figure 3. Several diseases at the bottom right are separated from the majority and we found that they are diseases spreading among both animals and humans, for example, Chronic Wasting Disease, Creutzfeldt-Jakob Syndrome, and Prion Disease.

\section{\bf Results}

From the 4,300 abstracts, the identified concepts were: 1,188 for Disease, 1,309 for Chemical, 822 for Gene, 322 for Species and 40 for SNP\&Mutation.These biomedical concepts form 21,521, 8,048, 5,042, and 161 unique relationships: Disease-Chemical, Disease-Gene, Disease-species, and Disease-SNP\&Mutation respectively. The most frequent Disease-Concept pairs can be seen in Table 1. We noticed that polyphenols, which are usually found in fruits and vegetables, have high co-existence with multiple neurodegenerative diseases. Polyphenols are well known for their function to reduce the risk of neurodegenerative disease [9]. Moreover, plenty of epidemiological studies reported the relation of intaking Omega-3 on Alzheimer’s Disease [10]. The appearance of Omega-3 in chemical-disease pairs in Table 1 further verified this. Olea europaea and Curcuma longa are the two top-ranked food-related species that are associated with multiple diseases. Diabetes Mellitus, Cardiovascular Diseases, and Obesity are the three top-ranked non-neurodegenerative diseases that appear in disease-disease pairs. From all the 4,300 abstracts, only 20 abstracts contain SNP\&Mutation-Disease pairs, which indicates that there might be a lack of research focusing on this area. We did not show these pairs in Table 1 as their occurrences are quite sparse. 

\begin{table}[ht!]
\vspace{3mm}
\caption{Most frequent entity-entity pairs extracted from Pubmed}
\begin{center}

\resizebox{\columnwidth}{!}{\centering \small \begin{tabular}{||c c c | c c c ||} 
\hline
 & Chemical-Disease Pair \hspace{2cm} &  &  & Gene-Disease Pair\hspace{3cm} &  \\ [0.5ex] 
 Chemical Name & Disease Name & Count & Gene Name & Disease Name & Count\\ [0.5ex]

 \hline\hline
 Polyphenols & Neurodegenerative Diseases & 175 & Abeta & Alzheimer Disease & 169\\ 
 \hline
 Lipids & Alzheimer Disease & 167 & tau & Alzheimer Disease & 110\\ 
 \hline
  Fatty Acids, Omega-3 & Alzheimer Disease & 131 & insulin & Alzheimer Disease & 108\\ 
 \hline
  Lipids & Neurodegenerative Diseases & 124 & Apo-E & Alzheimer Disease & 80\\ 
 \hline
   Polyphenols & Alzheimer Disease & 122 & Abeta & Neurodegenerative Diseases & 75\\ 
 \hline
   Polyphenols & Neoplasms & 108 & insulin & Diabetes Mellitus& 74\\ 
 \hline
 \hline
 & Species-Disease Pair  \hspace{2cm} &  &  & Disease-Disease Pair \hspace{3cm} &  \\ [0.5ex] 
 Species Name & Disease Name & Count & Disease Name & Disease Name & Count\\ [0.5ex]
  \hline
 \hline
 
  Olea europaea & Neurodegenerative Diseases & 37 & Alzheimer Disease & Diabetes Mellitus & 275\\ 
   \hline
  
    Olea europaea &  Neoplasms & 29 & Neurodegenerative Diseases & Diabetes Mellitus & 261\\ 
   \hline
    Olea europaea & Alzheimer Disease & 28 & Diabetes Mellitus & Cardiovascular Diseases & 212\\ 
   \hline
   Curcuma longa & Alzheimer Disease & 26 & Alzheimer Disease & Cardiovascular Diseases & 194\\ 
   \hline
    Curcuma longa & Neurodegenerative Diseases & 22 & Parkinson Disease  & Obesity & 146\\ 
   \hline   
     Curcuma longa & Neoplams & 20 & Neurodegenerative Diseases & Obesity & 135\\ 
   \hline     
\end{tabular}\par}
\end{center}
\end{table}

In the embedding space, we found the top-10 nearest neighbors of two representative neurodegenerative diseases: Alzheimer’s Disease and Parkinson’s Disease. Chemicals and food-related species that might be related to diet are highlighted in Table 2. We also included a more general concept, neurodegenerative diseases, which may refer to one or more diseases from literature.

\begin{table}[ht!]
\vspace{3mm}
\caption{Nearest neighbours of 3 diseases \label{data1}}
\begin{center}
\resizebox{\columnwidth}{!}{\centering \small \begin{tabular}{||c c c | c c c | c c c ||} 
 \hline
 & Alzheimer’s Disease \hspace{1cm} &  &  & Parkinson's Disease \hspace{1cm} &  &  & Neurodegenerative Diseases\hspace{1cm} & \\ [0.5ex] 
 Name & Distance & Type & Name & Distance & Type & Name & Distance &Type\\ [0.5ex]

 \hline\hline

 Oryctolagus cuniculus& 1.32 & Species & \cellcolor{blue!20} Amines & 1.65 & Chemical& Age & 1.52 & Gene\\ 
\hline
 AChE & 1.43 & Gene & rasagiline & 1.66 & Chemical& Endocannabinoids& 1.56 & Chemical\\ 
 
\hline 
 insulin receptor & 1.44 & Gene & Nicotine& 1.70 & Chemical& \cellcolor{blue!20} Polysaccharides& 1.56 & Chemical\\ 

\hline 
 \cellcolor{blue!20} Panax ginseng & 1.45 & Species & Dronabinol & 1.77 & Chemical& \cellcolor{blue!20} Allium sativum & 1.57 & Species\\ 

\hline 
 Malondialdehyde & 1.45 & Chemical & entacapone & 1.78 & Chemical& \cellcolor{blue!20} Sphingolipids & 1.57 & Chemical\\ 
 
 \hline 
 \cellcolor{blue!20} Zinc & 1.46 & Chemical & CB2 & 1.80 & Gene &PX clade & 1.58 & Species\\ 
 
  \hline 
 Abeta & 1.47 & Gene & Arrhythmias, Cardiac & 1.81 & Disease &\cellcolor{blue!20} Isoflavones & 1.58 & Chemical\\ 
 
   \hline 
 Fluorodeoxyglucose F18 & 1.47 & Chemical & \cellcolor{blue!20} Mucuna pruriens& 1.82 & Species & \cellcolor{blue!20} Thiamine & 1.58 & Chemical\\ 
 
    \hline 
 Clioquinol & 1.48 & Chemical & Uric Acid & 1.82 & Chemical &\cellcolor{blue!20} Agaricus bisporus & 1.58 & Species\\ 
    \hline 
 Silicon & 1.51 & Chemical & \cellcolor{blue!20}Cysteine & 1.83 & Chemical &\cellcolor{blue!20} Crocus sativus & 1.61 & Species\\ 
  \hline

\end{tabular}\par}
\end{center}
\end{table}

\section{\bf Conclusion}

In this study, we built a framework to construct and visualize a knowledge graph to link neurodegenerative diseases-related biomedical knowledge from PubMed. Specifically, we focused on relationships between neurodegenerative diseases and  food/diet. Our preliminary analysis indicated that the pipeline can be used to identify biomedical concepts that are semantically closed to each other as well as to reveal relationships between diet and diseases of interest. 

A breadth of possibilities exists to further improve this framework. For this paper, we only assumed there is some relation between concepts based on their co-occurrence. In the future, we could adopt natural language processing tools to extract relations between concepts to specify different relationships. Also, linking sparse knowledge from fast-growing literature would be beneficial for existing knowledge/information retrieval, and may promote uncovering of new knowledge. 
This framework is flexible and can be used for other applications such as drug repurposing,  therapeutic discovery, and clinical decision support for neurodegenerative diseases and other diseases. The knowledge graph constructed can facilitate researchers for data-driven knowledge discovery and new hypothesis generation.

\section*{\bf Acknowledgments}

This paper is partially supported by the National Institute of Health under award number RF1AG072799.

\bibliographystyle{apalike}
{\fontsize{10}{10}\selectfont

}
\end{document}